\documentclass{article}
\usepackage{amsmath}
\usepackage{array}
\usepackage{lipsum}
\usepackage{hyperref}
\usepackage{fancyvrb}
\usepackage{fvextra}
\usepackage{listings}
\usepackage{arxiv}
\usepackage[frozencache,cachedir=.]{minted}
\usepackage{xcolor}
\usepackage{longtable}
\usepackage{booktabs,tabularx,makecell}
\definecolor{LightGray}{gray}{0.9}

\title{Amadeus-Verbo Technical Report: The powerful Qwen2.5 family models trained in Portuguese.}
\author{William Alberto Cruz Castañeda, Marcellus Amadeus\\ \textbf{Amadeus AI}}
\date{}

\renewcommand{\headeright}{}
\renewcommand{\undertitle}{}
\renewcommand{\shorttitle}{}

\begin{document}
\maketitle

\begin{abstract}
This report introduces the experience of developing Amadeus Verbo, a family of large language models for Brazilian Portuguese. To handle diverse use cases, Amadeus Verbo includes base-tuned, merged, and instruction-tuned models in sizes of 0.5B, 1.5B, 3B, 7B, 14B, 32B, and 72B parameters. Thus, the main objective is to show how easy it is to fine-tune foundation models to democratize the open-source development of Brazilian Portuguese LLMs when data and resources are available. Amadeus-Verbo family models are all available at HuggingFace \href{https://huggingface.co/collections/amadeusai/amadeus-verbo-qwen25-67cf2e7aae69ce2b3bcdcfda}{here}.  
\end{abstract}

\section{Introduction}
Today’s most advanced large language models (LLMs) remain proprietary. But since the release of the Llama model series\cite{touvron2023llamaopenefficientfoundation} ignited interest within the open-source community to develop better LLMs. Moreover, Llama-3\cite{grattafiori2024llama3herdmodels} has emerged as the state-of-the-art open-weight model series, narrowing the performance gap with leading proprietary models and widely acknowledged as GPT-4–level.

Currently, an increasing number of LLMs are now pursuing advancements similar to those made by the GPT series from OpenAI. One of these models is Qwen\cite{bai2023qwentechnicalreport}, which has been released in an open-weight manner and progressed to Qwen1.5\cite{qwen1-5}, Qwen2\cite{yang2024qwen2technicalreport}, and Qwen2.5\cite{qwen2025qwen25technicalreport}, all grounded in the Transformer architecture, and trained using next-token prediction. The open-weight model series from Qwen2.5 includes sizes of 0.5B, 1.5B, 3B, 7B, 14B, 32B, and 72B parameters. 

In this technical report, we introduce the post-training language models\cite{tie2025surveyposttraininglargelanguage} Amadeus-Verbo family, which are based on the Qwen2.5 open-weight model series and are fine-tuned in Brazilian Portuguese texts. For Amadeus-Verbo, fine-tuning constitutes a cornerstone of adapting pre-trained LLMs to specialized tasks, refining their capabilities through targeted parameter adjustments. This process leverages labeled or task-specific datasets to optimize performance, bridging the gap between general-purpose pre-training and domain-specific requirements. Thus, this work explores fine-tuning using annotated datasets in Brazilian Portuguese with the following hands-on contributions:

\begin{enumerate}
    \item \textbf{Infrastructure}, which explains the hardware and software requirements;
    \item \textbf{Dataset}, which explains the characteristics and structure of the dataset used;  
    \item \textbf{Post-Training}, which explains how a supervised fine-tuning training process is implemented as well as presents other technical details such as spent hours, and cost;
    \item \textbf{Evaluation}, which explains the implemented evaluation methodology and the benchmarking used;
\end{enumerate}

\section{Infrastructure}
The Amadeus-Verbo family builds on pre-trained Qwen2.5 base and instruct series models and extends them to full-precision full-parameter fine-tuning models on bfloat16, including 0.5B, 1.5B, 3B, 7B, 14B, 32B, and 72B parameters. We maintain the Transformer-based decoder architecture of Qwen2.5 series models, we used all the Qwen2.5 base and instruction models implementation available in the HuggingFace ecosystem \footnote{https://huggingface.co/Qwen}, and the license type is kept according to the original Qwen2.5 model size.

For the 0.5B, 1.5B, 3B, 7B, and 14B models it was used an AWS computing instance p5.48xlarge with 8 NVIDIA H100 Tensor Core GPUs with Ubuntu Linux. On the other hand, for the 32B and 72B models, it was used an AWS computing instance p5e.48xlarge with 8 NVIDIA H200 Tensor Core GPUs with Ubuntu Linux. Regarding the software, we previously installed and tested the following frameworks: TRL\footnote{https://github.com/huggingface/trl}\footnote{https://huggingface.co/docs/trl/index}, Llama-Factory\footnote{https://github.com/hiyouga/LLaMA-Factory}, Swift\footnote{https://github.com/modelscope/ms-swift}, unsloth\footnote{https://unsloth.ai/}, and Liger-Kernel\footnote{https://github.com/linkedin/Liger-Kernel}. We finally chose to use Swift for the following reasons:
\begin{itemize}
    \item Easily implementation of distributed training with distributed data-parallel (DDP), \textit{device\_map} simple model parallelism, DeepSpeed ZeRO2/ZeRO3, FSDP, and other distributed training techniques;
    \item Diverse hardware support, compatible with CPU, RTX series, T4/V100, A10, A100, H100, H200, Ascend NPU, and MPS, among others.
    \item Support of pure text large models, covering the entire process from training to deployment.
\end{itemize}

\section{Dataset}
Supervised fine tuning (SFT) adapts pre-trained LLMs to specific tasks by leveraging task-specific labeled datasets. Thus, SFT adjusts model parameters using annotated data, yielding models that are both precise and contextually attuned while preserving broad generalization capabilities. The dataset contains about $\approx 600k$ instructions instruction examples designed to conduct instruction tuning for pre-trained language models.

\subsection{Preparation}
Creating a high-quality SFT dataset is an essential process for successful fine-tuning. A typical SFT dataset consists of an instruction and its corresponding instance. This pair of elements enables the LLM to discern task-specific patterns and generate relevant outputs. Also, performing an SFT dataset screening ensures that only high-quality instruction–instance pairs remain in the final dataset. Listing \ref{fig:dataset_snipped} shows a snipped example of the structure of a data instance of the dataset. Where the output data field describes the answer to the instruction, and the instruction data field describes the task the model should perform. Each of the instructions is unique.

\begin{listing}[!ht]
\begin{minted}[breaklines, breakafter=d, frame=lines, framesep=2mm, baselinestretch=1.2, bgcolor=LightGray, fontsize=\footnotesize]{json}
{
  "instruction": "Crie uma tarefa de classificação agrupando a lista de itens fornecida.",
  "output": "Classe 1: Maçãs, Laranjas\nClasse 2: Bananas, Morangos\nClasse 3: Abacaxi",
}
\end{minted}   
    \caption{Structure of a data instance of the dataset.}
    \label{fig:dataset_snipped}
\end{listing}

An extended structure of the dataset could contain two additional data fields: input and text as shown in Listing \ref{fig:extended_dataset_snipped}. The input data field is an optional context or input for the task and the text data field is the instruction, input-output formatted with the prompt template used for fine-tuning the model.

\begin{listing}[!ht]
\begin{minted}[breaklines, breakafter=d, frame=lines, framesep=2mm, baselinestretch=1.2, bgcolor=LightGray, fontsize=\footnotesize]{json}
{
  "instruction": "Crie uma tarefa de classificação agrupando a lista de itens fornecida.",
  "input": "Maçãs, laranjas, bananas, morangos, abacaxis",
  "output": "Classe 1: Maçãs, Laranjas\nClasse 2: Bananas, Morangos\nClasse 3: Abacaxis",
  "text": "Abaixo está uma instrução que descreve uma tarefa, combinada com uma entrada que fornece mais contexto. Escreva uma resposta que complete adequadamente a solicitação.\n\n### Instrução:\nCrie uma tarefa de classificação agrupando a lista de itens fornecida.\n\n### Entrada:\nMaçãs, laranjas, bananas, morangos, abacaxis\n\n### Resposta:\nClasse 1: Maçãs, Laranjas\nClasse 2: Bananas, Morangos\nClasse 3: Abacaxis",
}
\end{minted}   
    \caption{Extended structure of a data instance of the dataset.}
    \label{fig:extended_dataset_snipped}
\end{listing}

\section{Post-training}
Once the dataset is prepared, the fine-tuning process begins with obtaining the Qwen2.5 base and Instruct models. During the fine-tuning phase, the model’s parameters are adjusted, which aligns the model with the requirements of a given application. With all Qwen2.5 models, we implement the full-parameter fine-tuning (base and Instruct).

A full-parameter fine-tuning refers to the process of adjusting all parameters of the Qwen2.5 models (base and Instruct), in contrast to parameter-efficient methods such as LoRA or Prefix-tuning, which modify only a subset of parameters. Full-parameter tuning is often preferred for tasks requiring high precision, but it entails substantial computational overhead. Also, memory optimization techniques, including mixed precision training and activation checkpointing, help reduce memory demands, enabling large models to be fine-tuned on systems with limited hardware resources.

Using the Swift framework, our full-parameter supervised fine-tuning code implementations follow a standard PyTorch structure and use the libraries tied to the HuggingFace ecosystem (Transformers, Datasets, Tokenizers, Accelerate), which allow further reproducibility. The following Command Line Interface (CLI) structure was used for Swift:


\begin{Verbatim}[breaklines=true, breakanywhere=true]
#for Qwen2.5 base model
swift sft \
    --model Qwen/Qwen2.5-XB \
    --dataset '<used_dataset>' \
    --output_dir Amadeus-Verbo-BI-Qwen2.5-XB-PT-BR-Instruct

#for Qwen2.5 Instruct model
swift sft \
    --model Qwen/Qwen2.5-XB-Instruct \
    --dataset '<used_dataset>' \
    --output_dir Amadeus-Verbo-FI-Qwen2.5-XB-PT-BR-Instruct 

# CLI script obtained from:
# https://github.com/modelscope/ms-swift/blob/main/examples/train/base_to_chat/full.sh
\end{Verbatim}

The results of the SFT of Qwen2.5-XB Base models were the Amadeus-Verbo Base Instruct (the outcome of Qwen2.5 Base + fine-tuning) models. The SFT of Qwen2.5-XB-Instruct models were the Amadeus-Verbo fine-tuning Instruct (the outcome of Qwen2.5 Instruct + fine-tuning) models. Thus the nomenclature is as follows:

\begin{Verbatim}[breaklines=true, breakanywhere=true]
Amadeus-Verbo-BI-Qwen2.5-XB-PT-BR-Instruct
Amadeus-Verbo-FI-Qwen2.5-XB-PT-BR-Instruct
\end{Verbatim}

\subsection{Parameters} \label{subsection: Parameters}
All the Qwen2.5 models (base and Instruct) were fine-tuned with 78840 examples for two epochs with a maximum length of 8192 tokens for a single sample. To optimize learning, the following training parameter configurations were used:

\break

\begin{Verbatim}[breaklines=true, breakanywhere=true]
CUDA_VISIBLE_DEVICES= {number_of_CUDA_devices} \
swift sft \
    --model_id_or_path {Huggingface_model_name} \
    --dataset HF::{Huggingface_dataset_name} \
    --max_length=8192 \
    --sft_type full \
    --seed 42
    --batch_size 1 \
    --max_steps -1 \
    --max_grad_norm 1 \
    --lr_scheduler_type cosine \
    --num_train_epochs 2 \
    --learning_rate 1e-5 \
    --optim: adamw_torch \
    --adam_beta1: 0.9 \
    --adam_beta2: 0.95 \
    --adam_epsilon: 1e-8 \
    --warmup_ratio 0.05 \
    --weight_decay: 0.01 \
    --acc_strategy token \
    --evaluation_strategy steps \
    --save_safetensors True \
    --gradient_accumulation_steps {steps_number} \
    --gradient_checkpointing True \
    --output_dir {output_directory_path} \
    --eval_steps {number_evaluation_steps} \
    --save_steps {number_save_steps} \
    --report_to all \
\end{Verbatim}

As shown in the parameters configuration above, for all Qwen2.5 models, we use the eight GPUs available (CUDA\_VISIBLE\_DEVICES) to perform distributed training. The type of training was full parameter fine-tuning with a default seed of 42, batch size=1, and max steps=-1. To deal with overfitting, we apply a weight decay of 0.01, and the gradient norms are clipped to a maximum value of 1.0.

To control the step size during optimization, we use a cosine learning rate. We use the default value of the learning rate as 1e-5 for full parameters and warmup ratio=0.05. The Adam optimizer was used as a replacement optimizer for gradient descent and to decrease error rates during training.

We use $adam\_beta1=0.9$ and $adam\_beta2=0.95$ as parameters to calculate the gradient averages, and $adam\_epsilon=1e-8$ to improve numerical stability. We save a model checkpoint every 500 steps and evaluate them on the same number of steps. We also use Weights \& Biases for tracking all the experiments.

\subsection{Hardware Resources}
Concerning hardware resources, we reserve an AWS p5.48xlarge with 8 NVIDIA H100 GPUs. For base models fine-tuning, we reserve 28 days with a price of 21,144.00 USD, and another 28 days for Instruct models fine-tuning with a total price of 42,288.00 USD. During this period, we configure all the software environments and with those compute instances, we fine-tuned the 0.5B, 1.5B, 3B, 7B, and 14B base and Instruct models. Table \ref{tab:table_1} shows the time (in hours and days) consumed for each model.

\begin{table}[h]
\centering
\caption{Equivalent days and hours spent by Base and Instruct Qwen2.5 fine-tuned models.}
\begin{tabular}{lcccc}
\hline
\textbf{Base/Instruct Models} & \textbf{Days} & \textbf{Total days} & \textbf{Hours} & \textbf{Total hours}\\ \hline
Qwen2.5-0.5B & 2.02 / 2.02 & 4.04 & 48.57 / 48.57 & 97.14 \\
Qwen2.5-1.5B & 2.8 / 2.8 & 5.6 & 67.92 / 67.92 & 135.84\\
Qwen2.5-3B  & 3.7 / 3.7 & 7.4 & 89.55 / 89.55 & 179.1\\ 
Qwen2.5-7B & 4.5 / 4.5 & 9 & 109.50 / 109.50 & 219\\ 
Qwen2.5-14B & 6.6 / 6.6 & 13.2 & 157.55 / 157.55 & 315.1\\ \hline
\textbf{Total} &  & \textbf{39.24} &  & \textbf{946.18} \\ \hline
\end{tabular}
\label{tab:table_1}
\end{table}

We reserve an AWS p5e.48xlarge with 8 NVIDIA H200 GPUs. For base models fine-tuning, we reserve 28 days with a price of 23,257.00 USD, and another 28 days for Instrcut models fine-tuning with a total price of 46,514.00 USD. During this period, we configured all the software environments, and with those compute instances, we fine-tuned the 32B and 72B models. Table \ref{tab:table_2} shows the time (in hours and days) consumed for each model.

\begin{table}[h]
\centering
\caption{Equivalent days and hours spent by Base and Instruct Qwen2.5 fine-tuned models.}
\begin{tabular}{lcccc}
\hline
\textbf{Base/Instruct Models} & \textbf{Days} & \textbf{Total days} & \textbf{Hours} & \textbf{Total hours}\\ \hline
Qwen2.5-32B & 5.24 / 5.24 & 10.48 & 125.76 / 125.76 & 251.52\\
Qwen2.5-72B & 18.4 / 18.4 & 36.8 & 440.50 / 440.50 & 881\\ \hline
\textbf{Total} & & \textbf{47.28} && \textbf{1132.52}\\ \hline
\end{tabular}
\label{tab:table_2}
\end{table}

\section{Model Merging}
Merge a model is a technique that combines two or more LLMs into a single model. We implement this technique using the mergekit library\footnote{https://github.com/arcee-ai/mergekit}. We merge Qwen2.5-XB-Instruct models with the corresponding Amadeus-Verbo-BI-Qwen2.5-XB-PT-BR-Instruct models, resulting in the Amadeus Verbo Merge Instruct (MI) models. Thus the nomenclature is as follows:

\begin{Verbatim}[breaklines=true, breakanywhere=true]
Amadeus-Verbo-MI-Qwen2.5-XB-PT-BR-Instruct
\end{Verbatim}

To create our model, we use the Spherical Linear Interpolation (SLERP) merge method due to that maintains a constant rate of change and preserves the geometric properties of the spherical space in which the vectors reside.

\subsection{Configuration}
We install mergekit directly from the source as follows:

\begin{Verbatim}[breaklines=true, breakanywhere=true]
!git clone https://github.com/cg123/mergekit.git
!cd mergekit && pip install -q -e .
\end{Verbatim}

Listing \ref{fig:yaml_file} shows the YAML file created with the merge configuration where is specified the names of the merged models. This configuration is applied to every layer of both models. Note that we input a gradient of values for the interpolation factor t. The parameters for the self-attention and MLP layers will use different combinations of Qwen/Qwen2.5-XB-Instruct and amadeusai/Amadeus-Verbo-BI-Qwen2.5-XB-PT-BR-Instruct. The other layers are a 50/50 mixture of the two models. To merge the models we run the merge command with the following parameters, which will download the weights of all the models listed in the merge configuration and run the selected merge method:

\begin{Verbatim}[breaklines=true, breakanywhere=true]
# Merge models
!mergekit-yaml config.yaml merge --copy-tokenizer --allow-crimes --out-shard-size 1B --lazy-unpickle
\end{Verbatim}

\begin{listing}[!ht]
\centering
\begin{minted}[breaklines, breakafter=d, frame=lines, framesep=2mm, baselinestretch=1.2, bgcolor=LightGray, fontsize=\footnotesize]{yaml}
slices:
  - sources:
      - model: Qwen/Qwen2.5-XB-Instruct
        layer_range: [0, 24]
      - model: amadeusai/Amadeus-Verbo-BI-Qwen2.5-XB-PT-BR-Instruct
        layer_range: [0, 24]
merge_method: slerp
base_model: Qwen/Qwen2.5-XB-Instruct
parameters:
  t:
    - filter: self_attn
      value: [0, 0.5, 0.3, 0.7, 1]
    - filter: mlp
      value: [1, 0.5, 0.7, 0.3, 0]
    - value: 0.5
dtype: bfloat16
\end{minted}   
    \caption{Merge configuration in a yaml format.}
    \label{fig:yaml_file}
\end{listing}

\subsection{Hardware Resources}
To merge Qwen2.5-XB-Instruct models with the Amadeus-Verbo-BI-Qwen2.5-xB-PT-BR-Instruct corresponding models, we reserve an AWS p5.48xlarge instance with 8 NVIDIA H100 GPUs for 14 days with a total price of 11,149.00 USD. During this period we also configure all the software environments.

\section{Evaluation}
For the evaluation of the 0.5B, 1.5B, 3B, 7B, and 14B models it was used an AWS computing instance p4de.24xlarge with 8 NVIDIA A100 Tensor Core GPUs with Ubuntu Linux. We reserve these instances for 28 days with a total price of 10,226.00 USD. For the 32B and 72B models, it was used an AWS computing instance p5e.48xlarge with 8 NVIDIA H200 Tensor Core GPUs with Ubuntu Linux.

\break

The base-tuned, instruction-tuned, and merged models were evaluated on a natively Brazilian Portuguese implementation of the EleutherAI LM Evaluation Harness made available by \cite{open_pt_llm}. We install the framework for evaluating LLMs in Portuguese directly from source as follows:

\begin{Verbatim}[breaklines=true, breakanywhere=true]
git clone https://github.com/eduagarcia/lm-evaluation-harness-pt
cd lm-evaluation-harness-pt
pip install -e .
\end{Verbatim}

Table \ref{tab:table_3} shows the descriptions and main metrics of the evaluation suite with nine diverse sets of tasks that cover different capabilities. The evaluations primarily use few-shot examples (typically 3 to 25, depending on the task) to assess model performance in context.

\begin{table}[htb]
\centering
\caption{Task descriptions and main metrics of the Portuguese language model evaluation suite.}
\begin{tabularx}{\linewidth}{l X c}
\toprule
\textbf{Task} & \textbf{Description} & \textbf{Main metric}\\
\midrule
assin2\_rte & dataset for recognizing textual entailment in Portuguese, part of the ASSIN 2 shared task. & F1 Macro\\
assin2\_sts & dataset for semantic textual similarity in Portuguese, assessing model ability to determine semantic equivalence between sentences. & Pearson\\
bluex & reading comprehension dataset for Portuguese, testing the ability to understand and extract information from texts. & F1 Macro\\
enem & questions from the Brazilian high school national exam (ENEM), covering various subjects in multiple-choice format. & Accuracy\\
faquad\_nli & natural language inference dataset derived from the FaQuAD question-answering dataset for Portuguese. & F1 Macro\\
hatebr & dataset of Brazilian instagram comments annotated for offensive language and hate speech detection. & F1 Macro\\
hate\_speech & hierarchically labeled Portuguese hate speech dataset composed of tweets with binary annotations. & F1 Macro\\
tweetsentbr & corpus of tweets in Brazilian Portuguese annotated for sentiment analysis in three classes (positive, negative, neutral). & F1 Macro\\
oab\_exams & multiple-choice questions from the Brazilian bar exam, testing legal knowledge and reasoning. & Accuracy\\
\bottomrule
\end{tabularx}
\label{tab:table_3}
\end{table}

We evaluate base, instruct, and merge models in each task using all the GPUs available in the compute instance with the following code:

\begin{Verbatim}[breaklines=true, breakanywhere=true]
accelerate launch -m lm_eval \
    --model hf \
    --model_args pretrained=MODEL_ID,revision=main,tokenizer=NAME,device_map=auto,low_cpu_mem_usage=True\
    --tasks enem_challenge,bluex,oab_exams,assin2_rte,assin2_sts,faquad_nli,hatebr_offensive,portuguese_hate_speech,tweetsentbr\
    --batch_size auto \
    --output_path results/YOUR_MODEL_ID \
\end{Verbatim}

\subsection{Base-tuned Models}
Table \ref{tab:table_4} shows the comparison between 0.5B, 1.5B, 3B, 7B, 14B, 32B, and 72B Qwen2.5-XB-Instruct and Amadeus-Verbo-BI-Qwen2.5-XB-PT-BR-Instruct models, highlighting only the tasks in which Amadeus-Verbo-BI-Qwen2.5-XB-PT-BR-Instruct performed equal or better.

\begin{table}[htb]
\centering
\caption{Tasks where Amadeus-Verbo-BI-Qwen2.5-XB-PT-BR-Instruct models were equal or better performance.}
\begin{tabular}{lcc}
\hline
\textbf{Models} & \textbf{Tasks} & \textbf{Metric value}\\
\midrule
\makecell{Qwen2.5-1.5B-Instruct\\Amadeus-Verbo-BI-Qwen2.5-1.5B-PT-BR-Instruct} &  assin2\_rte  & 0.88 / 0.88\\ \hline
\makecell{Qwen2.5-7B-Instruct\\Amadeus-Verbo-BI-Qwen2.5-7B-PT-BR-Instruct} &  assin2\_sts  & 0.76 / 0.81\\ \hline
\makecell{Qwen2.5-14B-Instruct\\Amadeus-Verbo-BI-Qwen2.5-14B-PT-BR-Instruct} &  \makecell{assin2\_rte\\faquad\_nli}  & \makecell{0.95 / 0.95\\0.80 / 0.83}\\ \hline
\makecell{Qwen2.5-32B-Instruct\\Amadeus-Verbo-BI-Qwen2.5-32B-PT-BR-Instruct} &  \makecell{hate\_speech\\assin2\_sts\\faquad\_nli}  & \makecell{0.74 / 0.75\\0.80 / 0.83\\0.83 / 0.87}\\ \hline
\makecell{Qwen2.5-72B-Instruct\\Amadeus-Verbo-BI-Qwen2.5-72B-PT-BR-Instruct} &  \makecell{hate\_speech\\assin\_rte\\assin2\_sts\\faquad\_nli}  & \makecell{0.76 / 0.78\\0.95 / 0.95\\0.81 / 0.85\\0.84 / 0.88}\\
\hline
\end{tabular}
\label{tab:table_4}
\end{table}

\subsection{Instruction-tuned Models}
Table \ref{tab:table_5} shows the comparison between 0.5B, 1.5B, 3B, 7B, 14B, 32B, and 72B Qwen2.5-XB-Instruct and Amadeus-Verbo-FI-Qwen2.5-XB-PT-BR-Instruct models, highlighting only the tasks in which Amadeus-Verbo-FI-Qwen2.5-XB-PT-BR-Instruct performed equal or better.

\LTcapwidth=\textwidth
\begin{longtable}{lcc}
\caption{Tasks where Amadeus-Verbo-FI-Qwen2.5-XB-PT-BR-Instruct models were equal or better performance.}\\
\label{tab:table_5}\\
\hline
\textbf{Models} & \textbf{Tasks} & \textbf{Model value} \\
\hline
\endfirsthead
\multicolumn{3}{c}%
{\tablename\ \thetable\ -- \textit{Continued from previous page}} \\
\hline
\textbf{Models} & \textbf{Tasks} & \textbf{Model value} \\
\hline
\endhead
\hline ticolumn{3}{r}{\textit{Continued on next page}} \\
\endfoot
\hline
\endlastfoot
\makecell{Qwen2.5-0.5B-Instruct\\Amadeus-Verbo-FI-Qwen2.5-0.5B-PT-BR-Instruct} &  bluex  & 0.35 / 0.36\\ \hline
\makecell{Qwen2.5-1.5B-Instruct\\Amadeus-Verbo-FI-Qwen2.5-1.5B-PT-BR-Instruct} & \makecell{oab\_exams\\assin2\_rte\\assin2\_sts}  & \makecell{0.44 / 0.44\\0.88 / 0.88\\0.76 / 0.76}\\ \hline
\makecell{Qwen2.5-3B-Instruct\\Amadeus-Verbo-FI-Qwen2.5-3B-PT-BR-Instruct} & \makecell{oab\_exams\\hate\_speech\\assin2\_rte\\assin2\_sts\\faquad\_nli}  & \makecell{0.47 / 0.47\\0.68 / 0.70\\0.92 / 0.92\\0.79 / 0.80\\0.77 / 0.77}\\ \hline
\makecell{Qwen2.5-7B-Instruct\\Amadeus-Verbo-FI-Qwen2.5-7B-PT-BR-Instruct} & \makecell{bluex\\hate\_speech\\assin2\_rte\\assin2\_sts}  & \makecell{0.64 / 0.65\\0.72 / 0.74\\0.94 / 0.94\\0.76 / 0.78}\\ \hline
\makecell{Qwen2.5-14B-Instruct\\Amadeus-Verbo-FI-Qwen2.5-14B-PT-BR-Instruct} & \makecell{assin2\_rte\\faquad\_nli}  & \makecell{0.95 / 0.95\\0.80 / 0.81}\\ \hline
\makecell{Qwen2.5-32B-Instruct\\Amadeus-Verbo-FI-Qwen2.5-32B-PT-BR-Instruct} & \makecell{hatebr\\assin2\_rte\\assin2\_sts\\faquad\_nli}  & \makecell{0.92 / 0.92\\0.95 / 0.95\\0.80 / 0.80\\0.83 / 0.84}\\ \hline
\makecell{Qwen2.5-72B-Instruct\\Amadeus-Verbo-FI-Qwen2.5-72B-PT-BR-Instruct} & \makecell{bluex\\hatebr\\assin2\_rte\\assin2\_sts\\faquad\_nli}  & \makecell{0.79 / 0.79\\0.88 / 0.89\\0.95 / 0.95\\0.82 / 0.83\\0.81 / 0.82}\\
\hline
\end{longtable}

\subsection{Merged Models}
Table \ref{tab:table_6} shows the comparison between 0.5B, 1.5B, 3B, 7B, 14B, 32B, and 72B Qwen2.5-XB-Instruct and Amadeus-Verbo-MI-Qwen2.5-XB-PT-BR-Instruct models, highlighting only the tasks in which Amadeus-Verbo-MI-Qwen2.5-XB-PT-BR-Instruct performed equal or better.

\LTcapwidth=\textwidth
\begin{longtable}[h]{lcc}
\caption{Tasks where Amadeus-Verbo-MI-Qwen2.5-XB-PT-BR-Instruct models were equal or better performance.}\\
\label{tab:table_6}\\
\hline
\textbf{Models} & \textbf{Tasks} & \textbf{Model value} \\
\hline
\endfirsthead
\multicolumn{3}{c}%
{\tablename\ \thetable\ -- \textit{Continued from previous page}} \\
\hline
\textbf{Models} & \textbf{Tasks} & \textbf{Model value} \\
\hline
\endhead
\hline \multicolumn{3}{r}{\textit{Continued on next page}} \\
\endfoot
\hline
\endlastfoot
\makecell{Qwen2.5-0.5B-Instruct\\Amadeus-Verbo-MI-Qwen2.5-0.5B-PT-BR-Instruct} & tweetsentbr  & 0.42 / 0.45\\ \hline
\makecell{Qwen2.5-1.5B-Instruct\\Amadeus-Verbo-MI-Qwen2.5-1.5B-PT-BR-Instruct} & \makecell{bluex\\oab\_exams}  & \makecell{0.52 / 0.52\\0.44 / 0.44}\\ \hline
\makecell{Qwen2.5-3B-Instruct\\Amadeus-Verbo-MI-Qwen2.5-3B-PT-BR-Instruct} & \makecell{bluex\\oab\_exams\\hate\_speech\\assin2\_rte}  & \makecell{0.57 / 0.58\\0.47 / 0.48\\0.68 / 0.69\\0.92 / 0.93}\\ \hline
\makecell{Qwen2.5-7B-Instruct\\Amadeus-Verbo-MI-Qwen2.5-7B-PT-BR-Instruct} & \makecell{bluex\\oab\_exams\\hate\_speech\\assin2\_rte\\assin2\_sts}  & \makecell{0.64 / 0.65\\0.53 / 0.53\\0.72 / 0.72\\0.94 / 0.94\\0.76 / 0.81}\\ \hline
\makecell{Qwen2.5-14B-Instruct\\Amadeus-Verbo-MI-Qwen2.5-14B-PT-BR-Instruct} & \makecell{enem\\oab\_exams\\tweetsentbr\\assin2\_rte\\faquad\_nli}  & \makecell{0.81 / 0.81\\0.61 / 0.61\\0.73 / 0.75\\0.95 / 0.95\\0.80 / 0.82}\\ \hline
\makecell{Qwen2.5-32B-Instruct\\Amadeus-Verbo-MI-Qwen2.5-32B-PT-BR-Instruct} & \makecell{bluex\\oab\_exams\\hate\_speech\\assin2\_rte\\assin2\_sts\\faquad\_nli}  & \makecell{0.78 / 0.78\\0.63 / 0.64\\0.74 / 0.74\\0.95 / 0.95\\0.80 / 0.81\\0.83 / 0.85}\\ \hline
\makecell{Qwen2.5-72B-Instruct\\Amadeus-Verbo-MI-Qwen2.5-72B-PT-BR-Instruct} & \makecell{oab\_exams\\assin2\_rte\\assin2\_sts\\faquad\_nli}  & \makecell{0.67 / 0.68\\0.95 / 0.95\\0.82 / 0.83\\0.81 / 0.84}\\
\end{longtable}

\section{Conclusion}
This technical report aims to document the experience and insights gained during the fine-tuning of a complete family of language models for the Brazilian Portuguese. As a result, we have successfully created a series of instruction language models, the Amadeus-Verbo family. We find that our fine-tuned and merged models perform comparably to similarly sized original instructional language models on Brazilian language tasks. In the future, we will focus on advancing robust developments on Amadeus-Verbo pre-trained models. First, we will refine models by incorporating broader, more diverse, higher-quality Brazilian Portuguese data. Second, we are committed to improve the reasoning capabilities of all the Amadeus-Verbo models to contribute, and stimulate the open development of Brazilian Portuguese LLMs.

\bibliography{main}
\bibliographystyle{plain}

\end{document}